# From Skeletons to Semantics: Design and Deployment of a Hybrid Edge-Based Action Detection System for Public Safety

Ganen Sethupathy[1], Lalit Dumka[2], and Jan Schagen[1]
[1]Sopra Steria Germany, Hamburg, Germany
[2]Sopra Steria India, Noida, India

Corresponding author: Ganen Sethupathy (e-mail: ganen.sethupathy@soprasteria.com).

This work was conducted as part of internal research and development activities at Sopra Steria.

**ABSTRACT** Public spaces such as transport hubs, city centres, and event venues require timely and reliable detection of potentially violent behaviour to support public safety. While automated video analysis has made significant progress, practical deployment remains constrained by latency, privacy, and resource limitations, particularly under edge-computing conditions. This paper presents the design and demonstrator-based deployment of a hybrid edge-based action detection system that combines skeleton-based motion analysis with vision-language models for semantic scene interpretation. Skeleton-based processing enables continuous, privacy-aware monitoring with low computational overhead, while vision-language models provide contextual understanding and zero-shot reasoning capabilities for complex and previously unseen situations. Rather than proposing new recognition models, the contribution focuses on a system-level comparison of both paradigms under realistic edge constraints. The system is implemented on a GPU-enabled edge device and evaluated with respect to latency, resource usage, and operational trade-offs using a demonstrator-based setup. The results highlight the complementary strengths and limitations of motion-centric and semantic approaches and motivate a hybrid architecture that selectively augments fast skeleton-based detection with higher-level semantic reasoning. The presented system provides a practical foundation for privacy-aware, real-time video analysis in public safety applications.

**INDEX TERMS** Action recognition, Computer vision, Edge computing, Public safety, Video surveillance

## I. INTRODUCTION

Public spaces such as transport hubs, city centres, and event areas require effective support for public safety and situational awareness. A key challenge is the timely detection of violent behaviour and physical confrontations in public environments. Video camera systems are widely used in these environments, but the scale and complexity of modern scenes exceed the capacity of continuous human monitoring.

Automated video analysis can support early detection of critical events, but its deployment in public spaces is constrained by strict requirements on latency, privacy, and transparency. Cloud-based processing is often unsuitable for security-sensitive applications, which has increased interest in edge-based video analysis, where data is processed locally close to the camera. Recent work has demonstrated that real-time and privacy-compliant intelligent video analytics can be deployed on edge devices for public safety applications [1].

Several methods for skeleton-based action analysis have been proposed in the literature. However, many existing approaches rely on proprietary components or are not designed for practical deployment on real edge devices. At the same time, recent advances in edge hardware, especially GPU-enabled platforms, have substantially increased the computational capabilities available directly at the network edge. In parallel, vision-language models (VLMs) enable semantic scene understanding and explanation but remain challenging to deploy under real-time edge constraints.

However, there is still a lack of open-source, edge-deployable approaches that systematically compare motion-based skeleton analysis and semantic vision-language models and integrate both within a controlled, agent-based workflow for public safety applications.

In addition, the experimental evaluation of such systems is constrained by regulatory and ethical requirements, which often limit real-world testing in public environments and motivate controlled, demonstrator-based validation.

This contribution addresses these gaps by presenting an agent-based edge video analysis approach for public safety. The main contributions of this work are:

(1) a practical comparison of skeleton-based action detection and vision-language models under real-time edge constraints with respect to latency, resource usage, interpretability, and false alarm behaviour.

(2) an open-source, agent-based architecture for edge video analysis.

(3) a multi-stage skeleton pipeline combining YOLO-Pose, ByteTrack, MotionBERT, and ProtoGCN for real-time action classification from body keypoints; and

(4) an interactive demonstrator showing real-time feasibility on an edge device connected to a camera, with a web-based monitoring dashboard.

This paper does not aim to propose new action recognition models or to outperform state-of-the-art benchmarks. Instead, it provides a system-level comparison and deployment-oriented analysis of existing paradigms under real-time edge constraints, with a focus on practical trade-offs among latency, resource usage, and operational feasibility.

This contribution focuses on system design, practical comparison, and demonstrator-based validation rather than large-scale quantitative evaluation. It is intended as a foundation for further research and operational evaluations in security-relevant environments. Fig. 1 provides a high-level overview of the edge-based video analysis concept considered in this work.

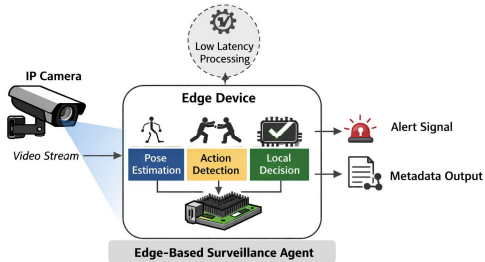

Fig. 1. High-level architecture of an edge-based video analysis system for public safety. Video streams from a camera are processed locally on an edge device, where perception and decision components operate without transmitting raw video data.

## II. System Overview and Design

### A. Edge Device

The system runs on an NVIDIA Jetson AGX Thor Developer Kit. This device is used to handle a high computational load directly at the edge.

We first tested the system on earlier Jetson-based embedded devices. These platforms work well for simple tasks. However, we observed limitations when running multiple real-time processes simultaneously. These limitations are primarily due to the Jetson Nano's limited memory and low computational power, which effectively prevent it from running most VLMs. While heavily quantised, minimal models may run, but they result in significantly degraded output quality. Similar limitations of edge computing devices, such as the Jetson Nano, under real-time deep learning workloads have been noted in prior work, where the execution of multiple concurrent inference tasks leads to performance bottlenecks and increased latency [2].

### B. Camera

A 5-megapixel RGB USB camera is connected directly to the edge device. The camera provides real-time video streams under different lighting conditions.

All video processing is performed locally on the edge device. Raw video data is not transmitted or stored outside the system. Short video segments are buffered only when needed.

### C. Agent Layer

The long-term goal of the system is to achieve an agent-based architecture that controls the perception and reasoning components. However, the agent layer is not fully implemented in the current prototype.

In the first development stage, the focus is on evaluating skeleton-based action detection and vision-language models as separate components. Model outputs are currently analysed directly, without automated agent-based orchestration. The agent layer is planned as future work to integrate model coordination, confidence handling, and human-in-the-loop workflows. Accordingly, the agent layer should be understood as a system-design concept rather than a fully operational component

### D. Software Architecture

The system is implemented as two independent FastAPI backend services sharing a common React-based frontend dashboard. This dual-backend design enables both pipelines to be tested independently and side-by-side on the same hardware.

The VLM backend operates on port 8080 and handles semantic video analysis through a vision-language model. The skeleton backend operates on port 8090 and performs action classification through pose estimation and graph convolutional networks. Both backends expose identical REST API and WebSocket interfaces for stream control, alert retrieval, statistics, and real-time video broadcast. Each backend follows a Producer-Consumer architecture. A StreamCapture component runs in a dedicated thread, reading frames from the camera or RTSP source via OpenCV and publishing them to multiple asyncio queues. Consumer tasks read from these queues to perform inference and broadcast frames to the UI. This separation ensures that frame acquisition remains decoupled from potentially slow inference, preventing frame drops under high computational load.

Data persistence is handled through SQLite with SQLAlchemy (async). Each analysis session is recorded with a start time and source URL. When the system detects a dangerous event, it creates an alert record containing the risk level, a text summary, the path to a saved video clip, and a thumbnail image.

Storage is organised hierarchically under a configurable root directory with subdirectories for raw recordings, alert clips, and optional inference-overlay recordings. This structure supports efficient post-hoc review and data management.

*E. Web-Based Monitoring Dashboard*

The frontend is a single-page React application built with Vite and styled with Tailwind CSS, designed as a surveillance-style monitoring dashboard. The application connects to either backend through a proxy layer that maps /api and /ws routes to the VLM backend (port 8080) and /skel-api and /skel-ws routes to the skeleton backend (port 8090). An operator can switch between backends at runtime without restarting the application.

The dashboard provides the following components:

1. Live Monitor: Displays the real-time video stream from the active backend. In skeleton mode, the stream includes an overlay with detected keypoints, skeleton connections, track IDs, and classification labels. In VLM mode, the stream shows the raw camera feed.
2. Alert Feed: A reverse-chronological list of detected events. Each alert shows its timestamp, risk level (colour-coded: red for DANGER, amber for WARNING, green for SAFE), and the text summary generated by the respective backend. Clicking an alert opens a modal with the saved video clip for review.
3. Real-Time Metrics Panel: Displays pipeline performance data.
4. Control Bar: Allows operators to enter an RTSP URL or select a local camera, start or stop streaming, and upload pre-recorded video files for offline analysis.

Fig. 2 shows the frontend dashboard during a live monitoring session.

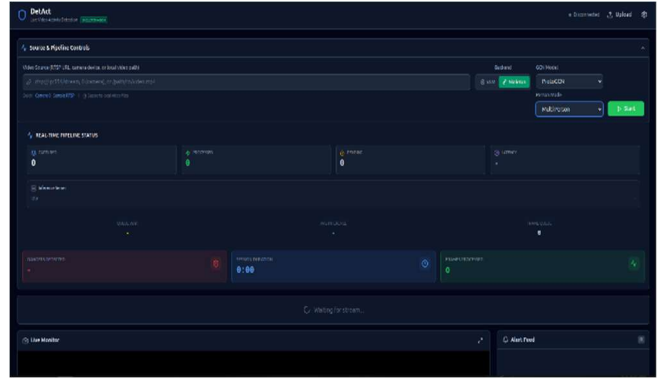
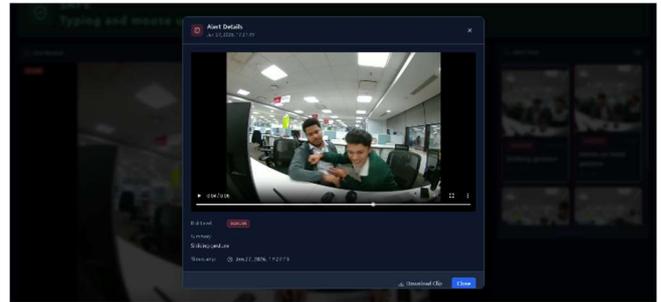

**Fig. 2. User Interface**

*F. System Configuration*

Both backends are configurable through environment variables using Pydantic Settings, allowing all operational parameters to be adjusted without code changes. Table 1 summarises the key configuration parameters.

TABLE I.
Key System Configuration Parameters

| Parameter Feature | VLM Backend (DETACT ) | Skeleton Backend (SKEL ) |
|---|---|---|
| Primary Model | Sehyo/Qwen3.5-35B-A3B-NVFP4 | ProtoGCN (NTU-60/120) |
| Inference Interval | chunk_duration_sec: 4.0s | YOLO runs at ~30 FPS |
| Clip Settings | clip_len: 100 | clip_stride: 30 |
| Sampling Rate | recent_fps: 6 (24 frames/chunk) | N/A |
| History Window | max_sec: 10.0s | fps: 1 | N/A |
| Resolution | target_max_dim: 720 | min_dim: 420 | Native Resolution |
| Generation Params | max_tokens: 10024 | temp: 0. temp: 0.4 | top_p: 0.6 | N/A |
| Detection Logic | dual_server_mode: True/False | yolo_confidence: 0.2 | iou: 0.9 |
| Tracking/Safety | N/A | pair_distance: 0px (optional) | max_persons: 100 |

## III. Skeleton-Based Action Detection

Skeleton-based action detection analyses human motion using body keypoints instead of raw image appearance. Rather than processing full video frames, the approach focuses on body posture and movement over time. This makes it well-suited for detecting physical violence in public spaces, where strong body movements and close interactions are key indicators.

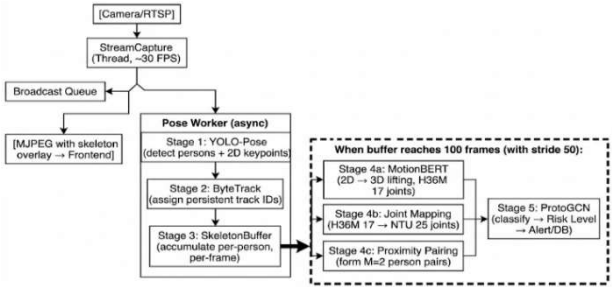

**Fig. 3. Data flow of the skeleton-based action detection pipeline**

Recent work has demonstrated that pose-driven approaches are effective for anomaly and safety-related activity detection in CCTV-like environments, further supporting the use of skeleton representations in surveillance scenarios [3]. Similar skeleton-based approaches have also been investigated in applied research projects for intelligent video surveillance in German cities, such as Mannheim [4].

*A. Pipeline Architecture*

The skeleton pipeline is implemented as a five-stage processing chain, executed as an asynchronous worker within the FastAPI application. Fig. 3 shows the complete pipeline data flow.

All deep learning models (YOLO-Pose, MotionBERT, ProtoGCN) are loaded once during application startup and shared across pipeline restarts. This avoids repeated model loading overhead and ensures consistent GPU memory allocation throughout the session.

*B. Pose Estimation (Stage 1)*

We use YOLOv26L-Pose [16] for multi-person 2D pose estimation. YOLO-Pose detects persons and extracts 17 COCO keypoints (nose, eyes, ears, shoulders, elbows, wrists, hips, knees, ankles) in a single forward pass. This bottom-up, single-stage architecture is critical for edge deployment: inference time remains approximately constant regardless of the number of persons in the scene, unlike top-down approaches such as MediaPipe [17], where cost scales linearly with person count.

We also evaluated MediaPipe Pose and OpenPose [18] during prototyping. MediaPipe provides 32 keypoints with finer hand and face detail but uses a top-down pipeline that degrades significantly in crowded scenes. OpenPose provides detailed pose estimation but requires higher computational resources. YOLO-Pose was selected for its balance of speed, accuracy, and scalability.

On the Jetson AGX Thor, YOLO26L-Pose achieves per-frame inference latency of approximately 21 ms in .pt format.

We exported the model to TensorRT format, which led to an improved average inference speed of 13.2ms per frame. The model internally resizes input to 640 pixels on the longest side, which explains the resolution-independent throughput. Recent survey work provides a comprehensive overview of modern deep learning–based human pose estimation techniques, covering both 2D and 3D approaches, and highlights current architectural trends, benchmarks, and open challenges in pose extraction research [5].

*C. Multi-Person Tracking (Stage 2)*

Persistent identity tracking is essential for accumulating per-person skeleton sequences over time. We use ByteTrack [23] to associate detections across frames.

ByteTrack's key insight is to associate every detection box, including low-confidence ones, rather than only high-score detections. This reduces identity switches and fragmented trajectories, which is important in our context because temporary occlusions (e.g., one person passing behind another) should not cause the skeleton buffer to lose track and restart accumulation.

*D. Skeleton Buffering (Stage 3)*

Each tracked person maintains an independent skeleton buffer (a sliding window of COCO 17 keypoints) that accumulates frame-by-frame. When a buffer reaches 100 frames (configurable via clip_len), it becomes eligible for emission. To avoid excessive computation, a stride of 30 frames is used: after emitting a clip, the buffer must accumulate at least 30 new frames before the next emission.

This design means a new classification is produced every ~1.0 seconds at 30 FPS (30 frames / 30 FPS), providing near-continuous monitoring while amortising the cost of 3D lifting and GCN inference. The buffer manager also handles person loss: if a tracked person disappears for more than 60 frames, their buffer is discarded. This prevents stale skeleton data from influencing classifications.

*E. 2D-to-3D Pose Lifting (Stage 4a)*

A critical challenge in using pretrained GCN models is that they expect 3D skeleton input (trained on NTU RGB+D [19][20], which uses Kinect depth sensors), while YOLO-Pose produces only 2D keypoints. We bridge this gap using MotionBERT [24], a transformer-based model that lifts temporal sequences of 2D poses to 3D.

MotionBERT uses a Dual-stream Spatio-temporal Transformer (DSTformer) architecture pretrained on Human3.6M to learn the underlying 3D structure from 2D observations. It accepts input of shape (T, 17, 3) — normalised 2D coordinates plus confidence per joint in H36M format — and outputs (T, 17, 3) 3D coordinates. If no MotionBERT checkpoint is available, the system falls back

*F. Joint Remapping Chain (Stage 4b)*

The GCN classifier (ProtoGCN) is trained on NTU RGB+D data, which uses 25 Kinect V2 joints, while MotionBERT outputs 17 Human3.6M joints, and YOLO-Pose detects 17 COCO joints. A three-stage remapping chain bridges these incompatibilities:

1. COCO 17 → H36M 17: Direct keypoint mapping with synthetic joint computation. Virtual joints not present in COCO (pelvis, spine, thorax, neck) are computed as midpoints of existing keypoints. For example, the pelvis is the midpoint of the left and right hips; the thorax is the midpoint of the left and right shoulders.

2. H36M 17 2D → H36M 17 3D: MotionBERT 2D-to-3D lifting (described above).

3. H36M 17 3D → NTU 25 3D: Mapping to the Kinect V2 skeleton layout. The 8 joints absent from H36M (left/right hand, left/right foot, hand tips, thumbs, spine_shoulder) are approximated from the nearest available H36M joint with small offsets.

Each mapping stage preserves temporal coherence across the clip. The confidence scores from COCO detections are propagated through the chain to maintain per-joint reliability information.

*G. Proximity-Based Interaction Pairing (Stage 4c)*

NTU RGB+D action classes include 26 mutual actions (A50–A60 in NTU-60, plus A61–A120 extensions) that involve two persons. The ProtoGCN model expects input with M=2 person slots. To handle this, the system pairs persons based on spatial proximity.

For each frame in the clip, the system computes skeleton centroids using the hip and shoulder keypoints (COCO indices 5, 6, 11, 12) with a confidence filter (threshold 0.3). Candidate interactions are generated as all pair combinations among active tracks, with optional distance gating (configurable; disabled by default in the current implementation). A person can therefore appear in multiple candidate pairs.

Paired persons are passed to the GCN as a two-person sample (M=2). In multi-person mode, single-person clips are also retained, with the second-person slot zero-padded. This increases interaction coverage while keeping pairing behaviour configurable.

*H. Action Classification with ProtoGCN (Stage 5)*

The final classification stage uses ProtoGCN [11]. ProtoGCN decomposes skeleton dynamics into learnable prototypes representing core motion patterns of action units. By contrasting prototype reconstructions, it identifies discriminative representations for similar actions.

The system also supports CTR-GCN [12] as an alternative backbone, selectable via UI. Before inference, the skeleton data is preprocessed following the official ProtoGCN pipeline:

1. PreNormalize3D: Centres the skeleton at the spine_mid joint (NTU joint 1) of person 0 at frame 0. Zero-padded frames are preserved. If person 1 has more valid frames than person 0, the persons are swapped.

2. UniformSampleDecode: Resamples the temporal dimension to exactly 100 frames using linear interpolation. This handles variable-length clips deterministically.

3. FormatGCNInput: Packs the data into the (nc, M, T, V, C) format expected by the model.

The model outputs a probability distribution over 60 or 120 action classes (depending on the NTU dataset variant). Risk classification is determined by aggregating probabilities across predefined class sets:

- DANGER classes include punching/slapping (A50), kicking (A51), and pushing (A52).

- WARNING classes include actions such as throwing (A7), staggering (A42), falling down (A43), and touching another person's pocket (A57).

- SAFE encompasses all remaining classes.

A DANGER alert is triggered when the cumulative probability of danger classes exceeds 0.3; a WARNING is triggered when warning class probability exceeds 0.5. These thresholds were empirically tuned during demonstrator testing.

Fig. 4 shows a simplified example of a physical confrontation between two persons, where skeleton representations capture motion patterns relevant for strike analysis, such as fast arm movements and reduced interpersonal distance.

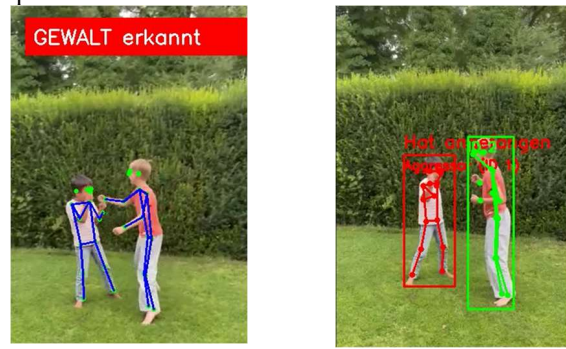

**Fig. 4. Skeleton-based representation of a physical interaction between two persons.**

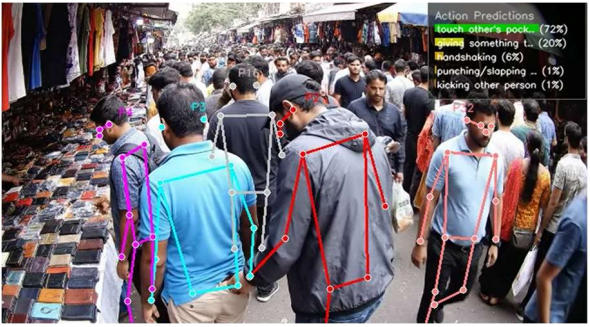

**Fig. 5. Illustrative example of skeleton-based action analysis in a crowded public space, showing body keypoints and motion patterns associated with a suspected pickpocketing interaction.**

The focus is on detecting events that may require rapid intervention, such as punches, kicks, or physical grappling between individuals.

Table II summarises representative results across two distinct modalities for action recognition. Skeleton-based GCN models (BlockGCN [11], CTR-GCN [12], InfoGCN [13]) report strong cross-subject performance on NTU-RGB+D—e.g., 93.1% on NTU-60 (X-Sub) [19] and 90.3% on NTU-120 (X-Sub) [20] for BlockGCN, with comparable ranges for CTR-GCN and InfoGCN [11–13]. In contrast, the VLM-based InternVideo2 [14] is evaluated on RGB video benchmarks and attains 92.1% (Kinetics-400 [21]) and 91.9% (Kinetics-600 [22]) under standard fine-tuning [14]. Because the NTU results use body-skeleton inputs and cross-subject protocols, whereas Kinetics figures reflect full-video RGB classification, the numbers are not directly comparable across rows; rather, they indicate that (i) skeleton-centric GCNs remain effective on pose-driven NTU settings, and (ii) VLM-based encoders provide competitive supervised accuracy on large-scale RGB video datasets.

Skeleton-based processing is suitable for edge deployment because it operates on compact keypoint data rather than high-resolution pixel information. This reduces computational load and supports real-time execution. In addition, the abstraction to body keypoints enables training and evaluation across diverse environments without relying on identifiable visual appearance, supporting a privacy-aware system design.

TABLE II.
Action Classification

| Model Architecture | Type | Dataset (split) | Top-1 Accuracy | Reference |
|---|---|---|---|---|
| BlockGCN | GCN | NTU-RGB+D 60 (X-Sub) | 93.1% | [11] |
| BlockGCN | GCN | NTU-RGB+D 120 (X-Sub) | 90.3% | [11] |
| CTR-GCN | GCN | NTU-RGB+D 60 (X-Sub) | 92.4% | [12] |
| InfoGCN | GCN | NTU-RGB+D 120 (X-Sub) | 89.8% | [13] |
| InternVideo2 | VLM | Kinetics-400 | 92.1% | [14] |
| InternVideo2 | VLM | Kinetics-600 | 91.9% | [14] |

### IV. Limitations of Skeleton + GCN

This system is fast; however, practical challenges remain. They suffer from inherent "context blindness." By abstracting humans into coordinate points, these systems discard critical visual information. They struggle to distinguish actions that are kinematically similar but contextually distinct—such as handing over a wallet (theft) versus shaking hands (greeting) or dancing vs fighting. Skeleton quality can degrade in crowded scenes, under partial occlusion, or in the presence of strong shadow effects. In some cases, shadows cast by persons are incorrectly detected as additional skeletons. Fig. 3 illustrates an example where a shadow is partially interpreted as a skeleton, introducing noise into the analysis. Such artefacts can often be identified by unstable keypoint trajectories or missing joint coherence across consecutive frames. Most critically, they are bound by closed-set supervision, incapable of detecting undefined or evolving threats.

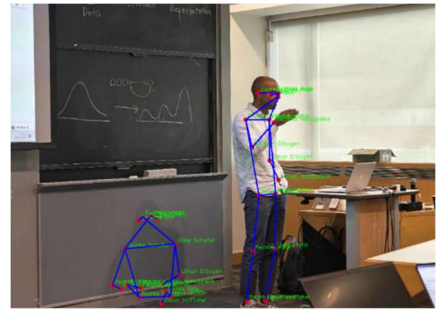

**Fig. 6. Illustration of limitations in skeleton-based pose estimation. Partial occlusion, shadows, and background structures can lead to incomplete or incorrect skeleton extraction in real-world environments.**

For basic violence detection, rule-based analysis of skeleton motion can identify strong arm or leg movements. However, distinguishing specific types of strikes (e.g., punching versus pushing) requires training action classes on sequences of skeleton data. This involves supervised learning on labelled skeleton trajectories to capture characteristic motion patterns of different physical interactions. This supervised learning requirement is consistent with the broader literature on skeleton-based action recognition, where learning from labelled skeleton sequences has been shown to be essential for distinguishing fine-grained motion patterns [6]

Some of the observed limitations, such as false detections caused by shadows or partial occlusions, are likely to be mitigated by combining observations from multiple cameras or by coordinating several sensing nodes as agents. In addition, hybrid approaches that combine skeleton-based detection as a fast first-stage filter with higher-level semantic reasoning can further improve robustness. These directions are discussed in Section V and evaluated as part of future work in Section VIII.

Overall, skeleton-based action detection provides an efficient and privacy-aware foundation for violence detection on edge devices. Its limited semantic understanding motivates the comparison with vision-language models and the investigation of combined approaches in the following sections.

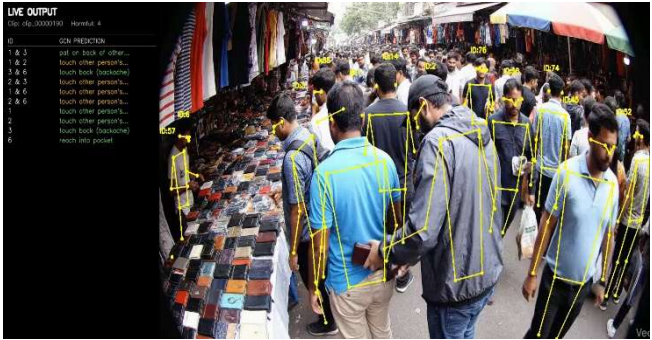

**Fig. 7. Illustration of limitations in skeleton-based Action classification.**

## V. Vision Language Models for Action Classification

The transition from traditional pose estimation to Vision Language Models (VLMs) marks a fundamental shift in surveillance technology, moving from analysing geometric motion to understanding semantic intent.

The following table illustrates the massive leap in performance enabled by VLMs compared to traditional ZSL methods.

TABLE III. COMPARISON OF VIDEO-STAR AND INTERNVIDEO2 ACROSS STANDARD ACTION-RECOGNITION BENCHMARKS (TOP1-ACC(%))

| Model | UCF-101 | HMDB-51 | K400 | K600 | Reference |
|---|---|---|---|---|---|
| Video-STAR-7B | 99.7 (B2N-HM) | 92.5 (CD) | 96.7 (B2N-HM) | 98.2 (CD) | [15] |
| InternVideo2 (Zero-Shot) | 89.5 (ZS) | 56.7 (ZS) | 73.1 (ZS) | 72.8 (ZS) | [14] |
| InternVideo2 (Fine-Tuned) | 97.3 (FT) | 80.7 (FT) | 92.1 (FT) | 91.9 (FT) | [14] |

ZS = Zero-Shot, FT = Fine-Tuned, B2N-HM = Base-to-Novel (harmonic mean), CD = Cross-Dataset.

Table III summarises the performance of Video-STAR [15] and InternVideo2 across four standard action-recognition benchmarks. The results show that modern vision–language models (VLMs) can achieve strong open-vocabulary and cross-dataset accuracy even without task-specific training, as seen in the zero-shot and base-to-novel evaluations. When fine-tuning is applied, performance improves consistently across all datasets, indicating that VLMs provide a robust foundation for both training-free and supervised action-recognition settings. These observations highlight that VLM-based systems are viable for scenarios where training data is limited, while still offering competitive gains when additional supervision is available.

### A. The Semantic Advantage of VLMs

Vision Language Models (such as Qwen3.5-VL, Llama-Vision, Gemma) overcome the limitations of skeleton-based systems by processing the "whole scene" rather than just the actors' kinematics. These capabilities have been demonstrated in recent large-scale VLMs designed to jointly process visual and textual signals across diverse tasks [7]. By integrating visual encoders with large language reasoning, VLMs offer two distinct advantages for public safety:

1. Object & Context Awareness: They can identify interactions with objects (e.g., a weapon vs. a phone) and analyse the surrounding environment to validate a threat.
2. Zero-Shot Reasoning: Unlike skeleton models that require retraining for new classes, VLMs can detect "unseen" anomalies simply through natural language prompting (e.g., "detect aggressive behaviour"), making them highly adaptable to dynamic real-world scenarios.

Fig. 8 illustrates this semantic capability by comparing two moments of the same sequence and the corresponding natural language explanations generated by the vision-language model.

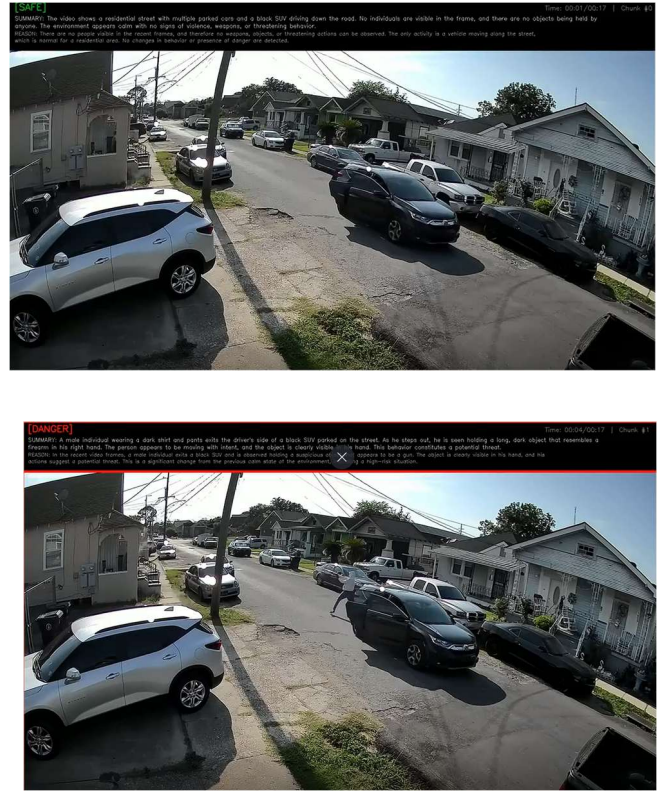

**Fig. 8 Illustrative comparison of vision-language model outputs for the same scene at two different time points.**

### B. Narrative Understanding Beyond Frame-Based Analysis

The system is designed to help the AI understand the story behind the video, rather than viewing events in isolation. Unlike standard methods that look at video frame-by-frame, our approach mimics how humans monitor situations using three key mechanisms:

1. Dual-Stream Sampling: Instead of looking at a single video clip, our system watches two timelines at once. It analyses a "Context Stream" (the past x seconds) to understand immediate scene context, while simultaneously watching an "Action Stream" (the last 4 seconds) to catch fast movements. This helps the model distinguish between safe actions (like running for a bus) and dangerous ones (like fleeing a crime scene).
   
   The context stream is sparse, 1FPS only and the action stream is at 6FPS. This is done to save the precious computation and achieve the real-time goal.

2. Recursive Narrative Inference: To reduce short-term forgetting, the system provides an optional memory loop. When enabled, the summary generated for the previous moment is fed back into the model as input for the current moment. This gives the VLM short-term memory, enabling it to track the progression of an event—such as identifying that a verbal argument is escalating into physical violence—rather than seeing it as a disconnected incident.

3. Zero-Shot Semantic Deployment: We utilise a large-scale VLM (Qwen3.5-VL) running on the edge itself. Instead of training the model on thousands of skeleton examples, we simply use natural language prompts to define what is dangerous. This allows the system to detect complex, evolving threats immediately without the need for retraining. However, the system can be tuned for specific deployments like indoors (factory, banks, etc) or outdoors (stadiums, crowded intersections, parks, etc). For e.g., the system will look for anomalies caused by vehicles (crashes, accidents) in an intersection, while it will not look for vehicles in the park.

## VI. System-Level Performance Evaluation (Edge Deployment)

To complement the architectural comparison, Table IV summarises representative system-level performance characteristics measured on the target edge device. The goal is not to benchmark recognition accuracy, but to illustrate latency, resource consumption, and operational trade-offs under real-time constraints. All measurements were obtained on the NVIDIA Jetson AGX Thor platform under comparable runtime conditions and represent indicative values observed during demonstrator operation rather than optimised benchmark results. The reported measurements are based on a limited set of demonstrator scenarios and are intended to illustrate system behaviour rather than provide statistically generalisable performance results.

TABLE IV. REPRESENTATIVE SYSTEM-LEVEL PERFORMANCE CHARACTERISTICS MEASURED ON THE TARGET EDGE DEVICE

| Metric | Skeleton-Based Layer | Vision-Language Layer |
| --- | --- | --- |
| Execution Mode | YOLO26x-pose (TensorRT) + MotionBERT + ProtoGCN, multi-person | Qwen3.5-35B-A3B (NVFP4), dual vLLM servers |
| Throughput | 41.9 eFPS | 1.34 eFPS |
| End-to-End Latency | 2.54 s (2.39 s buffer fill + 148 ms inference) | 5.49 s (chunk capture + inference) |
| Peak Unified Memory | 14.5 GB | 101.7 GB |
| GPU Power Draw | 36.3 W avg / 42.5 W peak† | 20.1 W avg / 34.1 W peak† |

*All measurements on Nvidia Jetson Thor (128 GB LPDDR5X unified memory, MAXN 130 W mode). Skeleton layer runs all models in-process; vision-language layer runs two external vLLM server processes hosting the quantised 35B model. Memory is reported as peak system memory (unified CPU–GPU on Jetson). GPU power draw is the iGPU rail only via pynvml (nvmlDeviceGetPowerUsage); total module power (CPU + GPU + DRAM + I/O) is higher but not instrumented here. Benchmarked on 9 videos totalling 458.5 s.*

*† The VLM GPU power reading (20.1 W) is measured from the benchmark orchestration process. The two vLLM server processes that perform the actual 35B model inference run as separate OS processes, and their GPU power draw is reflected in the same GPU rail but may not represent the full GPU load during inference bursts.*

The results confirm the complementary nature of both approaches. Skeleton-based processing provides stable real-time performance with low latency and moderate resource usage, making it suitable for continuous monitoring under crowded conditions. In contrast, vision-language-based reasoning introduces significantly higher computational overhead and latency but enables richer semantic interpretation of complex and previously unseen situations. These observations motivate the hybrid agent-based design discussed in the following section, in which fast motion-based detection selectively triggers higher-level semantic analysis.

## VII. Discussion and Practical Challenges

Deploying AI-based video analysis systems in public spaces raises practical challenges that go beyond model accuracy. Edge deployment limits computation, memory, and latency, especially in crowded scenes with many interacting people. Occlusions, shadows, changing lighting, and complex backgrounds can reduce the stability of both skeleton detection and semantic interpretation.

At the same time, edge hardware capabilities are evolving rapidly. Newer edge devices provide increased compute performance and memory, enabling more complex perception and reasoning workloads to be executed locally. This ongoing hardware progress reduces some current constraints and expands the design space for future edge-based safety systems. Recent work has surveyed optimisation strategies for deploying advanced AI models and autonomous reasoning agents on resource-constrained edge devices, emphasising the co-design of models and systems to make such workloads feasible under limited compute and memory budgets [8].

AI-based systems are also sensitive to noise and behaviour patterns. Sensor noise, motion blur, and small input changes can affect model outputs. In addition, repetitive movements or structured behaviour may trigger false alarms. While robustness can be improved through noise-aware training and cross-checking between different models, such effects cannot be fully avoided in real-world environments.

The proposed hybrid agent-based architecture helps mitigate several of these risks by separating fast motion-based detection from selective semantic reasoning. Skeleton-based analysis supports continuous monitoring under edge constraints, while vision-language models add contextual understanding and zero-shot flexibility. However, the coordination between these components introduces additional complexity. Incorrect triggering, delayed activation, or conflicting outputs may still lead to missed events or false alarms.

When agents are deployed on multiple edge devices, inter-device communication becomes a further challenge. While raw video data can remain local to each device, agents may exchange event metadata, confidence scores, or high-level alerts. This communication improves global situation awareness but introduces dependencies related to synchronisation, latency, reliability, and security. Designing communication protocols that are robust, privacy-aware, and resilient to partial failures remains an open challenge.

Deployment is further constrained by legal and regulatory conditions. In many regions, the legal basis for automated behaviour analysis in public spaces is still evolving. For example, recent revisions of police legislation in Germany have created explicit legal frameworks for the use of AI-assisted video analysis in public spaces. In the federal state of Hesse, legal changes enabled pilot deployments of intelligent video analysis in urban areas such as Frankfurt under defined legal and judicial conditions [9]. In other regions, including Berlin, further legislative adaptations are under discussion to expand the legal scope for video surveillance and automated analysis, although these changes remain subject to ongoing political and legal debate [10]. Even when models are open, run locally on edge devices, and do not transmit raw video externally, acceptance may depend on model transparency, governance, auditability, and the nature of inter-device information exchange.

As a result, system design involves clear trade-offs between performance, robustness, transparency, and regulatory acceptance. These trade-offs cannot be resolved only at design time. Instead, they must be explored through iterative testing and deployment in real operational environments, taking local regulations, organisational policies, and risk tolerance into account. This shows that practical suitability emerges through real-world use rather than purely technical optimisation.

In operational settings, the system should therefore be understood as a decision support tool that assists human operators, rather than as a fully autonomous decision-maker.

## VIII. Conclusion and Outlook

This work highlights a shift in public safety technology toward more versatile semantic interpretation. While previous methods were often limited to pre-defined object categories, modern vision-language models (VLM) can interpret complex events in a much broader, context-aware manner. By comparing skeleton-based action detection and vision-language models, we show that efficient motion-based approaches offer strong performance and privacy properties, while semantic models provide richer contextual understanding of complex situations.

The analysis indicates that neither approach is sufficient in isolation. Skeleton-based detection is limited by its lack of contextual awareness, whereas vision-language models introduce higher computational cost and operational complexity. Combining both enables a more balanced system design that better addresses the challenges of real-world public environments.

The major advantages of using a VLM-based system are:
- no major data collection overhead
- generalised solution with very little tailoring needed for each use case.

Looking forward, future work will focus on improving the feasibility and robustness of vision-language models on edge hardware. This includes tighter integration of motion-based cues with semantic reasoning, more efficient handling of multi-person interactions, and further optimisation of inference pipelines to reduce latency and resource usage. Future research will also explore more structured interaction modelling and systematic performance evaluation to better understand system behaviour under increasing scene complexity.

Rather than selecting one approach over the other, this work suggests that robust public safety systems emerge from their combination. The next development phase will focus on implementing the agent layer to coordinate motion-based detection and semantic reasoning autonomously on edge platforms, as well as evaluating scalability and decision consistency under realistic deployment conditions.

## IX. Acknowledgements

The authors acknowledge the use of generative artificial intelligence tools (specifically ChatGPT, OpenAI) solely for language editing and stylistic refinement of selected sections of the manuscript. No AI systems were used to generate scientific content, technical analyses, experimental results, or conclusions. All scientific content and interpretations are the sole responsibility of the authors.

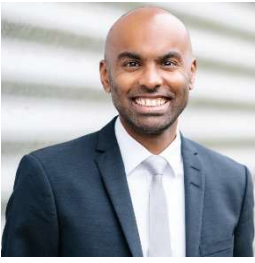
**Ganen Sethupathy** received the Ph.D. (Dr. rer. pol.) in economics from the University of Duisburg–Essen, Duisburg, Germany. His major field of study focused on explainable artificial intelligence, decision support systems, and their application in security-relevant and public-sector contexts. He is currently a Partner at Sopra Steria Germany, where he works on artificial intelligence, edge-based intelligent systems, and digital solutions for public safety and internal security. His professional experience includes applied research and large-scale system design at the intersection of AI, computer vision, and public-sector digitalization. His current research interests include edge AI, vision-based situational awareness, explainable AI, and deployment-oriented AI architectures for security-critical environments.

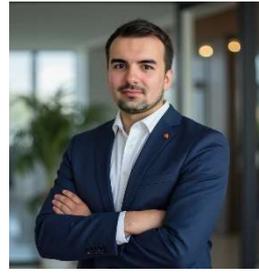
**Jan Schagen** received his B.S. degree in Business Administration and his M.S. degree in Economics from the University of Duisburg–Essen, Essen, Germany. In 2025, he also earned his Ph.D. (Dr. rer. pol.) there. From 2020 to 2024, he was a Research Associate at the Institute of Production and Industrial Information Management, University of Duisburg–Essen, Essen, Germany. His research focused on explainable artificial intelligence, decision support systems, ontologies, RAG systems, and their application in project management and public-sector contexts. He is currently a Senior Consultant at Sopra Steria Germany, where he works on artificial intelligence, information security, and digital solutions for public safety and internal security.

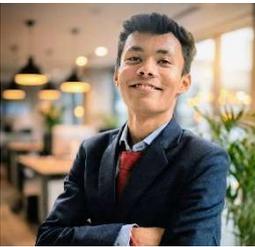
**Lalit Dumka** received his B.Tech. degree in Computer Science and Engineering from Graphic Era Hill University, India, in 2025. He is currently working in Applied Research & Machine Learning at Sopra Steria's AI & ML Center of Excellence, India. Previously, he served as a Research Intern at the Indian Institute of Technology (IIT) Roorkee, focusing on edge-deployed computer vision systems. His core research interests include Computer Vision, Generative AI, and Edge AI. Mr. Dumka is a recipient of the Best Paper Award at the IEEE CONECCT 2024 conference.